\documentclass[10pt, a4paper, oneside]{article}
\usepackage[hidelinks]{hyperref}
\usepackage{sci-paper}
\usepackage{times}
\usepackage{graphicx}
\usepackage{url}
\usepackage{ulem}
\usepackage{mathtools}
\usepackage{scalerel}
\usepackage{setspace}
\usepackage[strict]{changepage}
\usepackage{caption}
\usepackage[letterspace=-50]{microtype}
\usepackage{afterpage}
\usepackage{ragged2e}
\usepackage[ruled,vlined]{algorithm2e}
\usepackage{float}
\usepackage{hyperref}
\usepackage{titlesec}
 
\usepackage[margin=0cm, left=4.6cm, right=4.2cm, top=3.9cm, bottom=6.13cm, a4paper, headheight=0.5cm, headsep=0.5cm]{geometry}
\usepackage{fancyhdr}
\usepackage[format=plain, labelfont=it, textfont=it, justification=centering]{caption}
\usepackage{breakcites}
\usepackage{microtype}
\usepackage{geometry}
\usepackage{footmisc}

\hypersetup{
    colorlinks=true,
    linkcolor=blue,
    filecolor=magenta,      
    urlcolor=cyan,
}

 \titleformat*{\section}{\Large\bfseries}
 \titleformat*{\subsection}{\normalsize\bfseries}
 \graphicspath{{./img/}}
 \apptocmd{\frame}{}{\justifying}{}
 \newcommand\startingPage{1}
 \newcommand\paperauthor{{University of Pisa. }}

 \newcommand\papertitle{Sketches image analysis: Web image search engine using \newline LSH index and DNN InceptionV3}

 \header{\paperauthor \papertitle}

\begin{document}

 \title{{\fontsize{14pt}{14pt}\selectfont{\vspace*{-3mm}\papertitle\vspace*{-1mm}}}}

 \author{{\bfseries\fontsize{10pt}{10pt}\selectfont{Alessio Schiavo*, Filippo Minutella*, Mattia Daole*, Marsha Gómez Gómez \footnote[1]{Equal contribution. Listing order is random, all members cooperated providing important achievements.}}}\\
   {\fontsize{9pt}{12pt}\selectfont{University of Pisa, Department of Information Engineering, largo L. Lazzarino 1, 56122, Pisa, Italy
 }}
}

\maketitle

{\fontfamily{ptm}\selectfont
\begin{abstract}
{\fontsize{9pt}{9pt}\selectfont{\vspace*{-2mm}
The adoption of an appropriate approximate similarity search method is an essential prerequisite for developing a fast and efficient CBIR system, especially when dealing with large amount of data. In this study we implement a web image search engine on top of a Locality Sensitive Hashing (LSH) Index to allow fast similarity search on deep features. Specifically, we exploit transfer learning for deep features extraction from images. Firstly, we adopt InceptionV3 pretrained on ImageNet as features extractor, secondly, we try out several CNNs built on top of InceptionV3 as convolutional base fine-tuned on our dataset. In both of the previous cases we index the features extracted within our LSH index implementation so as to compare the retrieval performances with and without fine-tuning. In our approach we try out two different LSH implementations: the first one working with real number feature vectors and the second one with the binary transposed version of those vectors. Interestingly, we obtain the best performances when using the binary LSH, reaching almost the same result, in terms of mean average precision, obtained by performing sequential scan of the features, thus avoiding the bias introduced by the LSH index. Lastly, we carry out a performance analysis class by class in terms of recall against \textit{mAP} highlighting, as expected, a strong positive correlation between the two.}}
\end{abstract}}

{\fontfamily{ptm}\selectfont
\begin{keywords}
{\fontsize{9pt}{9pt}\selectfont{
InceptionV3 Indexing -- Similarity Search -- Classification -- Computer Vision -- Locality Similarity Hashing -- Convolutional Neural Network -- Image Analysis -- ImageNet}}
\end{keywords}}

\vspace{10mm}

\section{Introduction}\label{sec:1}

Our project consists in developing a web search engine for hand drawn sketches retrieval. A user can draw a sketch through the web interface which is used as a query to retrieve and return the most similar images. The aim is to build an efficient interactive CBIR system based on indexing deep features extracted by mean of a CNN: we need to provide the user with an answer which is both fast and relevant with respect to her information need. 

The project comprises five main phases. In Section \ref{sec:2}{} is about datasets selection and preparation. Section \ref{sec:3}{} concerns LSH index actual implementation. We implement two versions of the index, a first version indexing real numbers feature vectors based on the random projection method and a second version, known as SimHash LSH \cite{li2011theory}, working with vectors of binary features. Additionally, we implement an index free structure in order to store deep features sequentially. Section \ref{sec:4}{} presents aspects of carry out training, testing and performance evaluation of several CNNs architectures in order to investigate which is the best performing model to be used as deep features extractor out of images. In this phase we successfully exploit transfer learning techniques for our purpose. In Section \ref{sec:5}{} we carry our several tests on both of our LSH implementations so as to identify the best set of parameters. At last we perform a class by class performance analysis for the sake of investigating how the system behaves with each dataset class in terms of retrieval performances. Finally, Section \ref{sec:6}{} concludes the paper with a summary. 

\newpage

\section{Dataset}\label{sec:2}

For our project we use two different datasets serving different purposes. The main dataset is the Sketches dataset containing \(20,000\) labeled images belonging to 250 different classes (each class is represented by 80 image samples). The images represent black and white handmade sketches, are in png format and have a resolution of 1111x1111. 

The second dataset we adopt is taken from the \textit{MIRFLICKR-25000} \cite{Mirflickr} open evaluation project and it consists of \(25,000\) images downloaded from the social photography site Flickr through its public API. We exploit this dataset as a distractor for our CBIR system. We extract features from all samples and insert these into our index together with features extracted from sketches images in order to carry out some sort of robustness test: when querying our system by using a sketch image as query we would expect the system to return only sketches among top results as these are quite different with respect to \textit{MIRFLICKR} samples.   

\section{Index for Similarity Search}\label{sec:3}

At the core of our project there is Locality Sensitive Hashing algorithm. \cite{LSH} We need our system to provide the user not only with relevant results with respect to her information need but also in a relatively short amount of time. Since we need our system to be interactive, we cannot adopt exact similarity search methods as these do not scale at all, on the other hand, though approximate similarity algorithms do not guarantee to provide you the exact answer, they usually provide a good approximation and are faster and scalable. 

We adopt two different implementations of the LSH index. The main idea behind hash based indexing techniques is that given any two objects \(o_{1}\) and \(o_{2}\) from a dataset \(D\) and a hash function \(g()\), we want their hash value to collide with high probability \(Pr[g(o_{1}) = g(o_{2})]\) if the objects are similar, conversely, we want this probability to be low if the objects are not similar. In other words, we hash data points into buckets in such a way that data points near each other are located within the same bucket with high probability, while data points far from each other are likely to be in different buckets. 

\vspace{5mm}

\textbf{Real Valued Feature Vectors LSH.} Our first LSH implementation works with real valued feature vectors and works as follows. We build a set of \(L\) hash functions \(g_{1}, g_{2}, \dots, g_{L}\) with each \(g\) function obtained as the concatenation of \(k\) hash functions \(h_{1}, h_{2}, \dots, h_{k}\). Hence, given an object o we have \(g(o) = <h_{1}(o), h_{2}(o), \dots , h_{k}(o)>\). 

Each h hash function hi is obtained according to the formula \(h_{i}(o) = \lfloor \frac{(p*X_{i} + b_{i})}{w} \rfloor \), where 
\begin{itemize}
   \item \(w\) is the size of the segments in the projection vectors. (We use the value suggested by the authors which is \(4\)). \cite{LSHD}
   \item \(X_{i} = (x_{i,1}, x_{i,2}, \dots, x_{i,d})\) is a vector having the same dimensionality as the data points in our dataset and \(x_{i,j}\) is chosen from a Gaussian distribution
   \item \(b_{i}\) is a random scalar value in the range \([0, w]\)  
\end{itemize}

The following graphical geometric representation is a simple indexing example of three data points in a 3-dimensional space. Note that this corresponds to a single \(g()\) hash function obtained as the concatenation of three \(h\) hash functions, hence in this example we have \(L = 1\) and \(k = 3\). 

\begin{figure}[H]
   \includegraphics[width=13cm]{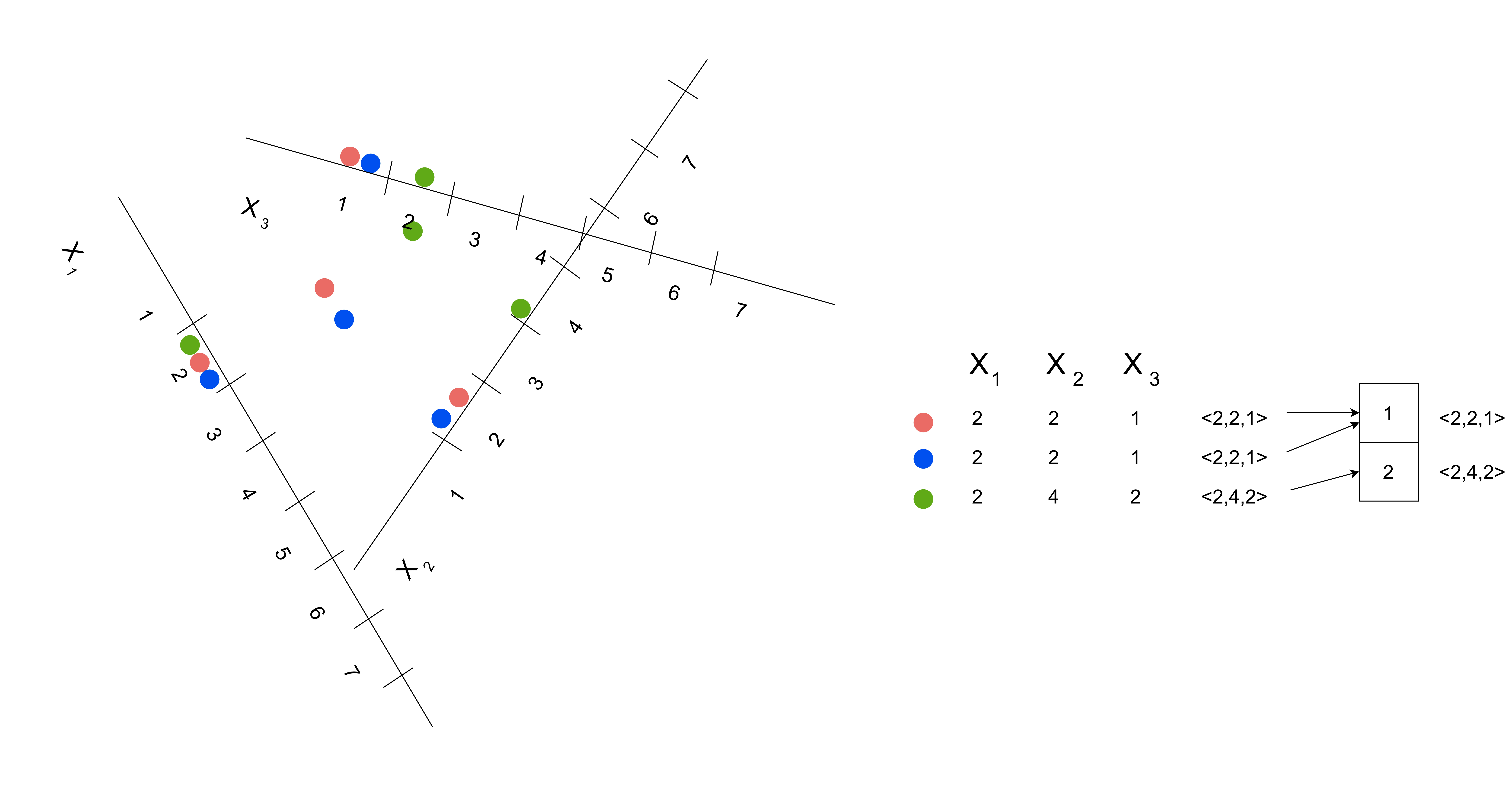}
   \centering
   \captionsetup{font=small}
   \caption{\fontsize{10pt}{11pt}\selectfont{\itshape{Projection base LSH}}}
   \label{fig:1}
\end{figure}

Any time a data points o is inserted in the index structure it is hashed and inserted into \(L\) buckets \(g_{1}(o), g_{2}(o),\dots, g_{L}(o)\). As far as the query execution is concerned, given a query object \(q\), we retrieve all data points from the \(L\) buckets \(g_{1}(q), g_{2}(q), \dots, g_{L}(q)\) and return the top k results ranked according to the distance function.

\textbf{Binary Valued Feature Vectors LSH.} We implement even a binary version of the LSH index. This method is known in literature as SimHash due to Moses Charikar \cite{binSimHash} and it is designed to approximate the cosine distance between vectors. The main idea behind this technique consists in choosing a random hyperplane (vector of coordinates randomly sampled from a normal distribution) and using the hyperplane to hash input vectors (data points). Given an input data point \(o\) and a hyperplane defined by \(r\), we have \(h(v) = sign(r \bullet v)\). In other words, each \(h\) hash function returns \(0\) or \(1\) as output value depending on whether the dot product \(r \bullet v\) is greater or equal than \(0\) (in that case \(v\) hashes to \(1\)) or not (\(v\) hashes to \(0\)). Each possible choice of \(r\) defines a single \(h\), and the hash function \(g\) is obtained as the concatenation of \(L\) hash functions \(h\). Hence, given an object \(o\) we have \(g(o) = <h_{1}(o), h_{2}(o), \dots, h_{k}(o)>\). If two objects \(o_{1}\) and \(o_{2}\) are similar, it is very likely that they will hash to the same value as they will be positioned similarly with respect to the \(K\) random hyperplanes defined.

\section{DNN used for features extraction}\label{sec:4}

We adopt as convolutional base for our CNNs models InceptionV3 \cite{InceptionV3} pre-trained on ImageNet, so as to exploit transfer learning from general purpose ImageNet images to sketches images. Specifically, as a first step, we evaluate the mean average precision \textit{mAP} obtained when indexing deep features extracted from our images by mean of the InceptionV3 without fine-tuning the network on top of the Sketches dataset. 

The \textit{mAP} value obtained this way serves as a baseline value: we want to understand if and by which margin we are able to improve performances when fine-tuning the CNN on top of our dataset. 

Additionally, we exploit several different networks variants by removing the fully connected part and by adding ourselves new custom sets of densely connected layers. 

\begin{figure}[H]
   \includegraphics[width=13cm]{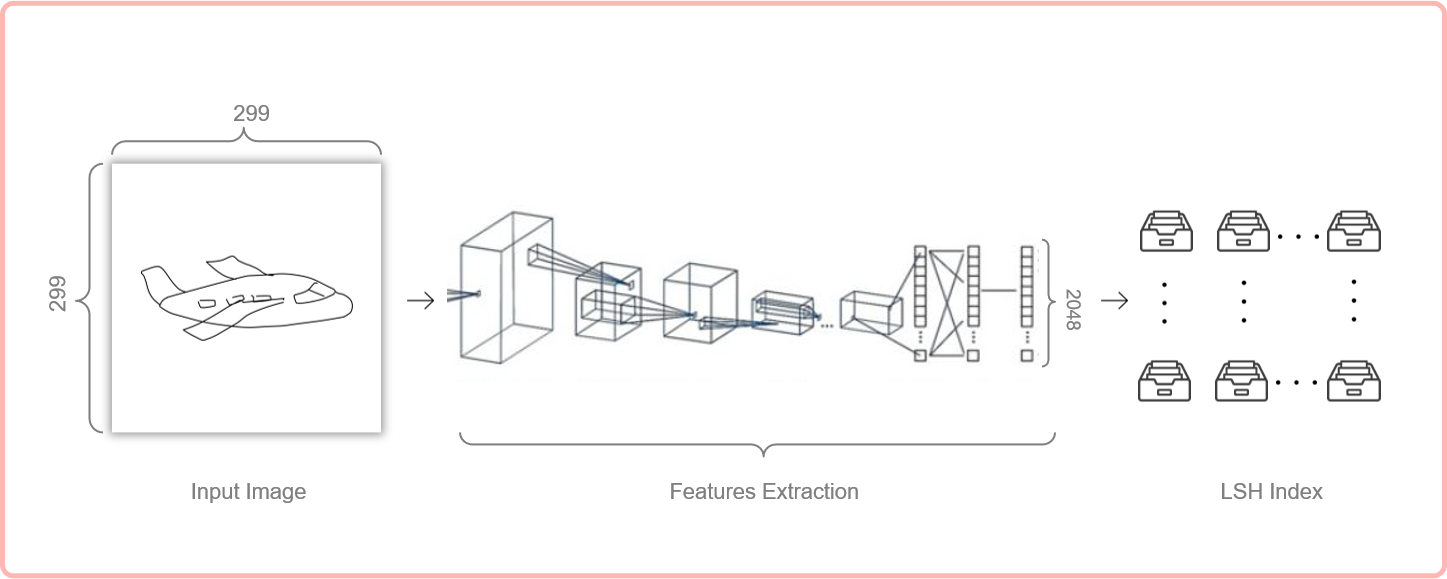}
   \centering
   \captionsetup{font=small}
   \caption{\fontsize{10pt}{11pt}\selectfont{\itshape{Feature extraction architecture}}}
   \label{fig:2}
\end{figure}

\subsection{Fine tuning strategy}

Throughout models training, we adopt two different fine-tuning strategies so as to investigate which is the best configuration for our specific problem and dataset. 

\begin{itemize}
   \item \textbf{2 Fine-tuning:} unfreeze the last two convolutional blocks and jointly fine-tune these with the custom fully connected part 
   \item \textbf{All Fine-tuning:} weights are initialized as ImageNet trained, but further train the entire network on our dataset 
\end{itemize}

For each fine-tuning strategy adopted, we performed the following steps:

\begin{enumerate}
   \item Add the custom network on top of an already-trained convolutional base network. 
   \item Freeze the convolutional base. 
   \item Train the fully connected added part. 
   \item Unfreeze some convolutional blocks 
   \item Jointly train both the unfrozen conv blocks and the fully connected added part. 
\end{enumerate}

\begin{figure}[H]
   \includegraphics[width=5cm]{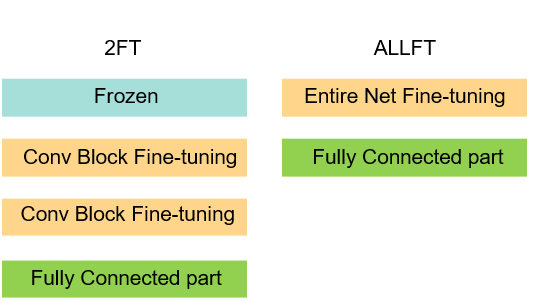}
   \centering
   \captionsetup{font=small}
   \caption{\fontsize{10pt}{11pt}\selectfont{\itshape{Fine tuning strategy}}}
   \label{fig:3}
\end{figure}

\subsection{Networks variants}

We build and train five different network architectures. We obtain four models by varying from time to time the structure of the fully connected added part. Moreover, we train a siamese network using one of the previously trained models as core CNN. We briefly report the structures of these models ordered by increasing model complexity. 

\begin{itemize}
   \item \textbf{Model Nr.1.} ImageNet pretrained Inception V3 as convolutional base on top of which we add a Global spatial average pooling layer followed by two fully connected layers. The first one having \(1024\) neurons and the second one having \(250\) neurons. The model has about \(24\) millions parameters. 
   \item \textbf{Model Nr.2.} ImageNet pretrained Inception V3 as convolutional base on top of which we add a Global spatial average pooling layer followed by two pairs of layers plus the output layer. Each pair made up of a dropout layer (with \(0.5\) as drop factor) and a fully connected layer having \(2048\) neurons. Finally, the output layer follows with \(250\) neurons. The model has about \(31\) millions parameters. 
   \item \textbf{Model Nr.3.} Siamese Network built using model nr.2 as core CNN and adding a lambda layer in order to compute the absolute difference between the encodings obtained from the two network inputs. The model has about 34 millions parameters. 
   \item \textbf{Model Nr.4.} Same structure as model nr.2 but with one more pair of dropout \(+ 2048\) fully connected layer before the output layer. The model has about \(35\) millions parameters.
   \item \textbf{Model Nr.5.} ImageNet pretrained Inception V3 as convolutional base on top of which we add a Global spatial average pooling layer followed by two pairs of layers plus the output layer. Each pair made up of a dropout layer (with \(0.5\) as drop factor) and a fully connected layer having \(4096\) neurons. Finally, the output layer follows with \(250\) neurons. The model has about \(48\) millions parameters. 
\end{itemize}

\subsection{Best Model Choice}

After training, we test each model in terms of mean average precision \textit{mAP}. Specifically, for each of the \(250\) classes in the sketches dataset, we hold out \(20\) out of 80 class samples as test set so as to use one image from the test set as query object. The \textit{mAP} is computed on a set of 250 queries, one per class, using cosine similarity as ranking measure. In the graph below we report model test \textit{mAP} (on the x axis) as a function of model complexity in terms of number of parameters (on the y axis). 

\begin{figure}[H]
   \includegraphics[width=13cm]{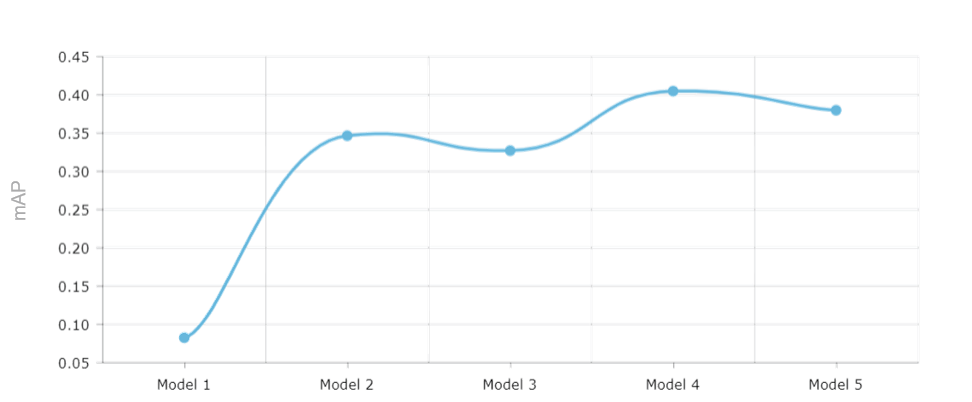}
   \centering
   \captionsetup{font=small}
   \caption{\fontsize{10pt}{11pt}\selectfont{\itshape{Test \textit{mAP} vs Model Complexity}}}
   \label{fig:4}
\end{figure}

As you can notice the best performing model in terms of \textit{mAP}, which is the best metric for \textit{CBIR systems evaluation}, is the Model number 4. Notice that, in this phase of the project, all models have been tested by performing sequential scan of the deep features so as to avoid the additional bias introduced by the LSH index approximation.  

\newpage

\section{LSH parameters tuning}\label{sec:5}

In LSH it is extremely important to properly tune the parameters in order to discover the best configuration in terms of trade-off between retrieval accuracy and efficiency: while the retrieval efficiency of the system improves the accuracy decreases accordingly since the approximation is tougher. As system retrieval accuracy metric we adopt test mean average precision \textit{mAP} (the same used for choosing the best network architecture). As far as system retrieval efficiency is concerned, we use Improvement In Efficiency \textit{IE} which is a metric computed as the ratio between the query cost without using the index (sequential scan of deep features) and the query cost when using LSH index. The cost is measured in terms of number of distance computations needed to answer a query.

LSH index needs to be properly tuned in both the real valued and binary cases: the optimal choice best suiting our system for the pair of parameters \((L,K)\) needs to be investigated. Increasing the number of projections axis (\(K\), number of hash functions \(h\)) has the effect of better separating dissimilar objects: the larger \(K\) the lower the probability of having two objects falling within the same bucket. Conversely, fixed \(K\), if we increase \(L\), the number of \(g\) hash functions, we increase the collision probability. We carry out several tests varying each time \(L\) and/or \(k\).

Moreover, in each test, beside computing \textit{mAP} and IE, we compute some useful statistics in order to derive better insights on the impact of LSH parameters on the system. Specifically, we measured: 

\begin{itemize}
   \item \textbf{Average Bucket Purity.} This is a pure number whose purpose is to provide a measure of \textit{“bucket purity”}: the larger is the number of samples belonging to the majority class within the bucket with respect to the total number of elements being in the bucket, the purest the bucket is. The average bucket purity is computed as the average of the ratios between the number of elements in the majority class and the total number of elements in each bucket.
   \item \textbf{Standard Deviation of Average Bucket Purity.} This is the standard deviation of the average bucket purity distribution. 
   \item \textbf{Number of Buckets.} This is simply the total number of distinct buckets making up the index. 
   \item \textbf{Number of Items.} This is the total number of objects contained within the index. 
   \item \textbf{Average number of elements per bucket.}
   \item \textbf{Standard Deviation of Average number of elements per bucket.} 
\end{itemize}

In the following table we show results obtained in correspondence of each real valued LSH index parameters configuration tried.

\begin{figure}[H]
   \includegraphics[width=13cm]{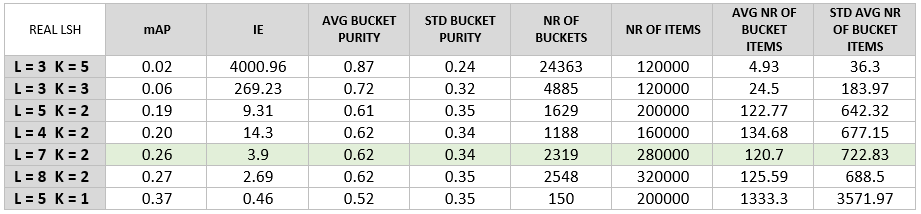}
   \centering
   \captionsetup{font=small}
   \label{fig:table1}
\end{figure}

\textbf{Observations.} Analyzing the numbers reported in the table you can notice that increasing the value of \(k\), the number of hash functions h, while the Improvement in Efficiency IE improves of three orders of magnitude (from \(0.46\) for \(k=1\) up to \(4,000\) for \(k=5\)), the \textit{mAP} of the system mutates in the opposite way (from the best value \(37\%\) for \(k=1\) down to \(0.02\) for \(k=5\)). Moreover, as k increases, the number of buckets increases by two orders of magnitude (from 150 buckets when k=1 up to \(24,636\) buckets when \(k=5\)) and the Average Bucket Purity increases (from \(52\%\) when \(k=1\) up to \(k=5\) when \(k=5\)) accordingly. These experimental results are aligned with what one would expect: increasing the number of projection axis, the probability that any two objects fall within different buckets increases. Hence, the number of buckets increases, the average number of items within each bucket decreases and as a consequence the Average Bucket Purity increases: it is as if you were implementing a more fine-grained clustering on data, thus having an higher probability of having samples belonging to the same class within each cluster. 

As far as parameter \(L\) is concerned, the number of g hash functions, which also corresponds to the number of index replicas you have, fixed \(K\), as \(L\) increases, while \textit{mAP} increases, \textit{IE} decreases (indeed the larger is \(L\) , the larger is the total number of items and the larger is the number of distance computations needed to answer to a query). Furthermore, being \(L\) the number of index replicas, if the index is held in main memory as it is the case in our implementation, it cannot exceed a certain value depending on the size of the data to be indexed and the amount of main memory available for the system. 

As highlighted in green within the table, in our specific use case, the parameters configuration yielding the best trade-off between system efficiency and accuracy is \(L = 7\) and \(K = 2\). With this configuration we obtain an \textit{IE} of \(4\) meaning that we have an improvement in retrieval cost by a factor of 4 and at the same time we do not loose too much in terms of \textit{mAP} (\(26\%\) against 40\% obtained when performing sequential scan without index). 

In the following table we show results obtained in correspondence of each binary valued LSH index parameters configuration tried.

\begin{figure}[H]
   \includegraphics[width=13cm]{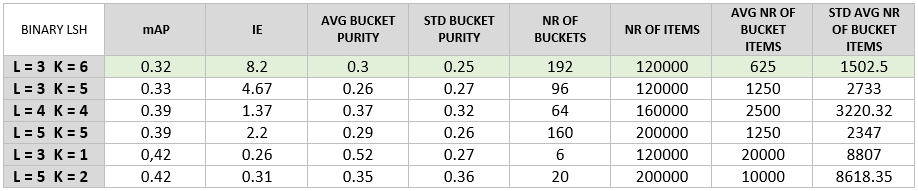}
   \centering
   \captionsetup{font=small}
   \label{fig:table2}
\end{figure}

\textbf{Observations.} As far as L and K parameters is concerned, similar observations as the real valued LSH index case. It is interesting to compare the best real valued LSH index result against the best binary valued LSH index in order to highlight the differences. The two results are reported in the table below.

As you can notice, in the Binary LSH case, we reach better performances both in terms of system efficiency with an IE of 8.2 against the 3.9 of the Real LSH and in terms of system accuracy with a \textit{mAP} of 32\% against the 26\% of the Real LSH. It is interesting to observe that the number of buckets in the binary case is much smaller (by an order of magnitude) than the real case (192 against 2319). Additionally, it can be noticed that the Average Bucket Purity value is smaller in the Real LSH case, indeed the average number of bucket items the buckets is about six times larger. Hence, in the binary case we have much less buckets each containing a larger number of objects, belonging to more classes: you would expect to obtain a worse \textit{mAP} value but it is not the case, since, as we will point out in the next paragraph, in the Sketches dataset there are often very similar samples belonging to different classes.

\begin{figure}[H]
   \includegraphics[width=13cm]{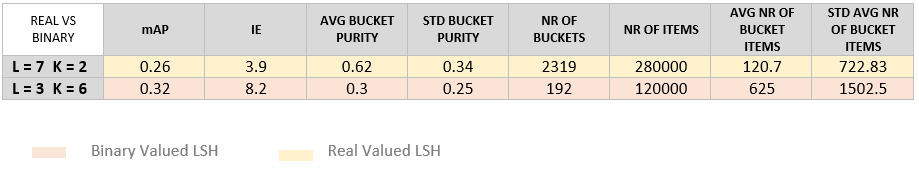}
   \centering
   \captionsetup{font=small}
   \label{fig:table3}
\end{figure}

\newpage

\section{Class by class analysis}\label{sec:6}

During the development of the project, when testing the system with a set of random queries, we noticed that the \textit{mAP} score obtained from time to time varied largely: when using a sample belonging to certain classes as query the \textit{mAP} score obtained was quite good while it was very bad in correspondence of some other samples belonging to other classes. Therefore, we thought of investigating in order to gain insights on specific classes by carrying out a class by class analysis. 

Specifically, for each individual class in the Sketches dataset, we computed both the \textit{mAP} and the CNN model Recall in such a way to be able to carry out a correlation analysis between the two. 

\vspace{5mm}

As far as the class \textit{mAP} computation is concerned, we proceeded as follows. 

\vspace{5mm}

\BlankLine{}

\IncMargin{1em}
\begin{algorithm}[H]
\SetKwData{Left}{left}\SetKwData{This}{this}\SetKwData{Up}{up}
\SetKwFunction{Union}{Union}\SetKwFunction{FindCompress}{FindCompress}
\SetKwInOut{Input}{input}\SetKwInOut{Output}{output}
\BlankLine{}
\emph{For each element class (\(c_i\)) in array class (\(C\))}\;
\ForEach{\(c_i \in C\)}{
   \emph{For each class sample (\(s_j\)) in array element class (\(c_i\))}\;
   \ForEach{\(s_j \in c_i\)}{
      execute\_query\( (s_j) \) \;
      \(AP \leftarrow \) compute\_AP\( (s_j) \) \;
   }
   \(mAP \leftarrow \) compute\_mAP\( (c_i) \) \;
   \(max \leftarrow \) compute\_max\( (AP, c_i) \) \;
   \(min \leftarrow \) compute\_min\( (AP, c_i) \) \;
   \(range \leftarrow \) compute\_range\( (AP, c_i) \) \;
   \(std \leftarrow \) compute\_std\( (AP, c_i) \) \;
   show\_best\_sketch\( (c_i) \) \;
   show\_worst\_sketch\( (c_i) \) \;
}
\caption{Class by class analysis process}\label{class_class}
\end{algorithm}\DecMargin{1em}
   
\BlankLine{} 

\vspace{10mm}

For each class, iteratively query the Sketches dataset, using each time one object the class as query to retrieve all the others. Compute the Average Precision for each query. Then, for each class, compute the class \textit{mAP}, the minimum and maximum \textit{mAP} values, the difference between the maximum \textit{mAP} and the minimum \textit{mAP} and the \textit{mAP} standard deviation. Finally, return the Sketch in correspondence of which the maximum \textit{mAP} value has been obtained and the one for which the minimum \textit{mAP} has been obtained. 

\vspace{5mm}

As output from this algorithm we obtained, for each class, a dictionary as the one that follows.

\begin{figure}[H]
   \includegraphics[width=7cm]{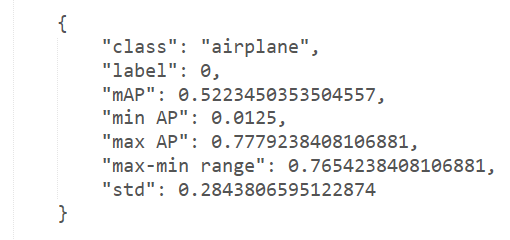}
   \centering
   \captionsetup{font=small}
   \caption{\fontsize{10pt}{11pt}\selectfont{\itshape{Output Class by Class algorithm}}}
   \label{fig:5}
\end{figure} 

\newpage

Beside this we computed the CNN model Recall score for each of the \(250\) Sketches dataset classes: what is the percentage of samples belonging to each class correctly recognized by our DL model. Then we carried out a correlation analysis between model class Recall and class \textit{mAP}, the scatter plot resulting is reported below.  

\begin{figure}[H]
   \includegraphics[width=7cm]{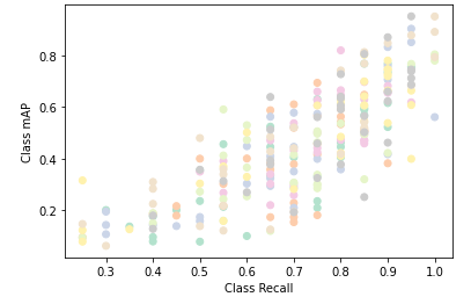}
   \centering
   \captionsetup{font=small}
   \caption{\fontsize{10pt}{11pt}\selectfont{\itshape{Recall and mAP by class}}}
   \label{fig:6}
\end{figure}

We obtained a correlation coefficient of \(0.79\) highlighting, as expected, a significant positive correlation between class Recall and class \textit{mAP}. 

We conclude by reporting two examples: a first one showing data obtained for one of the well-recognized classes and a second one showing data obtained in correspondence of one of the bad-recognized classes. 

Well-recognized class example:

\begin{figure}[H]
   \includegraphics[width=10cm]{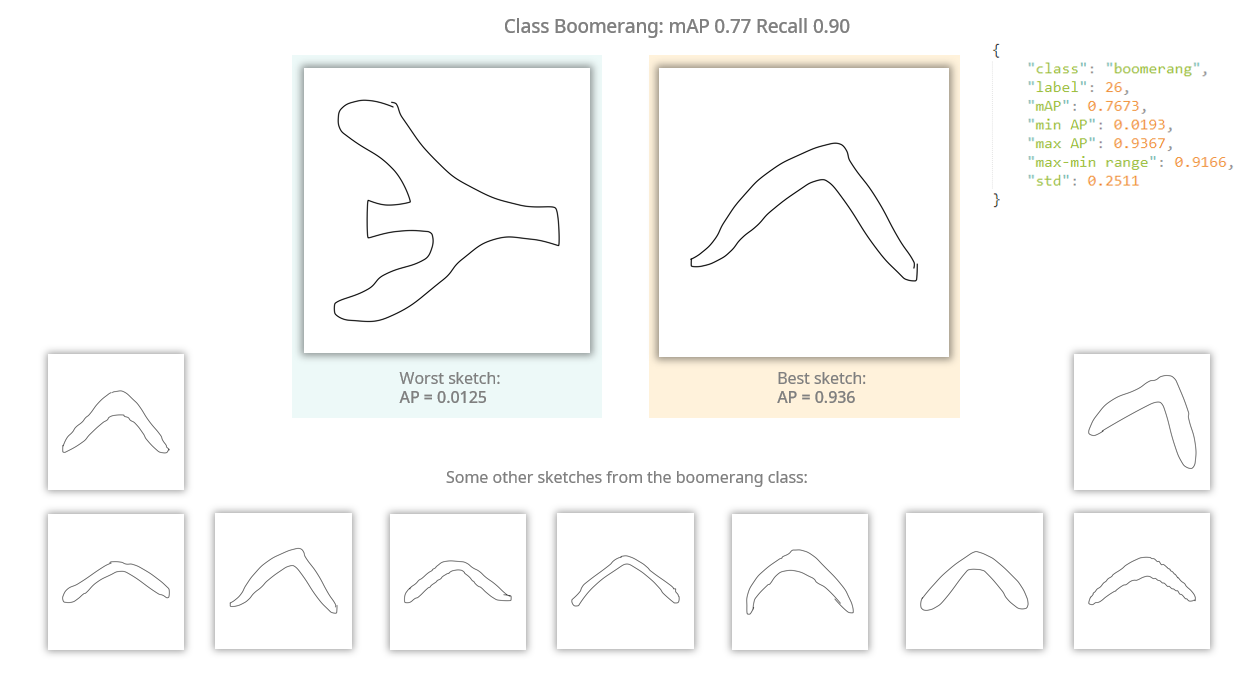}
   \centering
   \captionsetup{font=small}
   \caption{\fontsize{10pt}{11pt}\selectfont{\itshape{Well recognized class example}}}
   \label{fig:7}
\end{figure}

As it can be noticed, it this case the system provides the user with mostly relevant results expect when the input sketch is particularly badly drawn. Indeed, boomerang sketches are very similar one with respect to the others and the class Recall value is 90

\begin{figure}[H]
   \includegraphics[width=12cm]{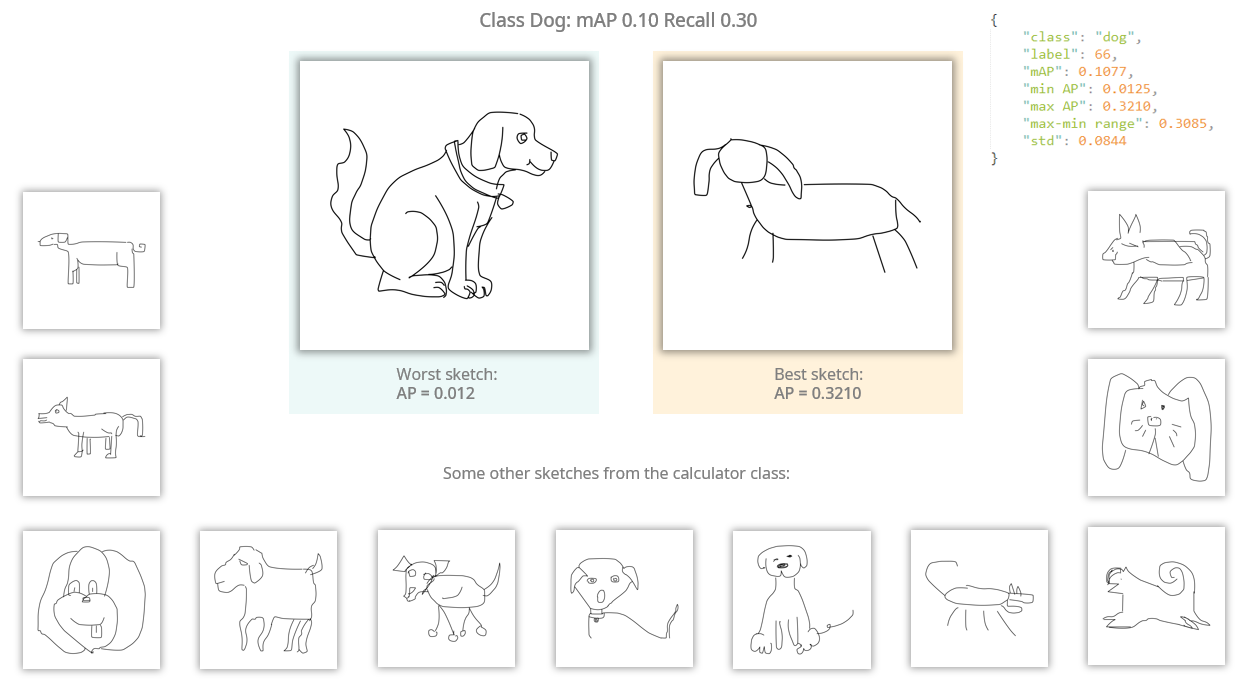}
   \centering
   \captionsetup{font=small}
   \caption{\fontsize{10pt}{11pt}\selectfont{\itshape{Bad recognized class example}}}
   \label{fig:8}
\end{figure}

The Dog class provides interesting results: if you look at the best and the worst sketch according to the Average Precision value obtained, these are reversed with respect to the judgment you would state according to your sense of aesthetics. This is due to the fact that, if you observe the other samples from the Dog class, these are all quite badly drawn, thus the only good looking dog sketch is an outlier yielding different features. Additionally, the class samples are all quite different one from the others, causing the low Recall and mAP values of respectively 30\% and 10\%. 

\vspace{10mm}

\section{Conclusion and Future Works}\label{sec:7}

Combining deep features extraction, exploiting transfer learning, and the approximate similarity search method LSH, we succeeded in building a CBIR system which is both efficient and accurate. Other indexing approaches could have been used for this purpose, e.g., scalar quantization \cite{amato2020large}. We developed a web application on top of the system so as to provide users with a friendly graphical interface through which they can draw sketches in input as queries. Most of the time the system provides fast and relevant answers to user needs.  

Video search engines, such as \cite{Visione2019, VisioneMDPI, Visione2021} developed by the AIMH Lab \cite{aimh2020}, would benefit from sketches image analysis. Integrating the propose approach with them is a future work.

\section*{Acknowledgments}

This work is the result of the student project made in the course ``Multimedia Information Retrieval and Computer Vision'' (Prof. Giuseppe Amato, Claudio Gennaro and Fabrizio Falchi) for the Master Degree in "Artificial Intelligence and Data Engineering" of the University of Pisa.

\newpage

\bibliography{Student-Paper}

\begin{thebibliography}{10}

\bibitem{aimh2020}
N.~Aloia, A.~Giuseppe, B.~Valentina, B.~Filippo, B.~Paolo, C.~Fabio,
  C.~Vittore, C.~Luca, C.~Cesare, C.~Silvia, E.~Andrea, F.~Fabrizio,
  G.~Claudio, L.~Gabriele, M.~F. Valerio, M.~Carlo, M.~Nicola, M.~Daniele,
  M.~Alessio, M.~Alejandro, N.~Alessandro, P.~Aandrea, P.~Nicolò, R.~Fauto,
  S.~Pasquale, S.~Fabrizio, T.~Costantino, T.~Luca, V.~Lucia, and V.~Claudio.
\newblock Aimh research activities 2020.
\newblock Technical Report 413891, Consiglio Nazionale delle Ricerche, 2020.

\bibitem{Visione2019}
G.~Amato, P.~Bolettieri, F.~Carrara, F.~Debole, F.~Falchi, C.~Gennaro,
  L.~Vadicamo, and C.~Vairo.
\newblock Visione at vbs2019.
\newblock In I.~Kompatsiaris, B.~Huet, V.~Mezaris, C.~Gurrin, W.-H. Cheng, and
  S.~Vrochidis, editors, {\em MultiMedia Modeling}, pages 591--596, Cham, 2019.
  Springer International Publishing.

\bibitem{VisioneMDPI}
G.~Amato, P.~Bolettieri, F.~Carrara, F.~Debole, F.~Falchi, C.~Gennaro,
  L.~Vadicamo, and C.~Vairo.
\newblock The visione video search system: Exploiting off-the-shelf text search
  engines for large-scale video retrieval.
\newblock {\em Journal of Imaging}, 7(5), 2021.

\bibitem{Visione2021}
G.~Amato, P.~Bolettieri, F.~Falchi, C.~Gennaro, N.~Messina, L.~Vadicamo, and
  C.~Vairo.
\newblock Visione at video browser showdown 2021.
\newblock In J.~Loko{\v{c}}, T.~Skopal, K.~Schoeffmann, V.~Mezaris, X.~Li,
  S.~Vrochidis, and I.~Patras, editors, {\em MultiMedia Modeling}, pages
  473--478, Cham, 2021. Springer International Publishing.

\bibitem{amato2020large}
G.~Amato, F.~Carrara, F.~Falchi, C.~Gennaro, and L.~Vadicamo.
\newblock Large-scale instance-level image retrieval.
\newblock {\em Information Processing \& Management}, 57(6):102100, 2020.

\bibitem{binSimHash}
M.~S. Charikar.
\newblock Similarity estimation techniques from rounding algorithms.
\newblock In {\em Proceedings of the Thiry-Fourth Annual ACM Symposium on
  Theory of Computing}, STOC '02, page 380–388, New York, NY, USA, 2002.
  Association for Computing Machinery.

\bibitem{LSHD}
M.~Datar, N.~Immorlica, P.~Indyk, and V.~S. Mirrokni.
\newblock Locality-sensitive hashing scheme based on p-stable distributions.
\newblock In {\em Proceedings of the Twentieth Annual Symposium on
  Computational Geometry}, SCG '04, page 253–262, New York, NY, USA, 2004.
  Association for Computing Machinery.

\bibitem{LSH}
A.~Gionis, P.~Indyk, and R.~Motwani.
\newblock Similarity search in high dimensions via hashing.
\newblock In {\em Proceedings of the 25th International Conference on Very
  Large Data Bases}, VLDB '99, page 518–529, San Francisco, CA, USA, 1999.
  Morgan Kaufmann Publishers Inc.

\bibitem{Mirflickr}
M.~J. Huiskes and M.~S. Lew.
\newblock The mir flickr retrieval evaluation.
\newblock In {\em Proceedings of the 1st ACM International Conference on
  Multimedia Information Retrieval}, MIR '08, page 39–43, New York, NY, USA,
  2008. Association for Computing Machinery.

\bibitem{li2011theory}
P.~Li and A.~C. K{\"o}nig.
\newblock Theory and applications of b-bit minwise hashing.
\newblock {\em Communications of the ACM}, 54(8):101--109, 2011.

\bibitem{InceptionV3}
C.~Szegedy, V.~Vanhoucke, S.~Ioffe, J.~Shlens, and Z.~Wojna.
\newblock Rethinking the inception architecture for computer vision.
\newblock {\em CoRR}, abs/1512.00567, 2015.

\end{thebibliography}
\bibliographystyle{abbrv}

\end{document}